\documentclass[letterpaper, 10 pt, conference]{ieeeconf}  

\usepackage{graphics}
\usepackage{caption}
\usepackage{subcaption}

\usepackage{graphicx}    
\usepackage{array}       
\usepackage{booktabs}    
\usepackage{tabularx}    
\usepackage{rotating}    
\IEEEoverridecommandlockouts                              
\usepackage[utf8]{inputenc}
\usepackage[T1]{fontenc}

\usepackage[table,xcdraw]{xcolor}
\usepackage{amssymb}
\usepackage{amsmath}
\usepackage{multirow}
\usepackage{multicol}

\definecolor{rblue}{rgb}{0,0.5,1}
\definecolor{awesome}{rgb}{1.0, 0.13, 0.32}
\definecolor{hollywoodcerise}{rgb}{0.96, 0.0, 0.63}
\definecolor{lasallegreen}{rgb}{0.03, 0.47, 0.19}
\definecolor{hanpurple}{rgb}{0.32, 0.09, 0.98}
\definecolor{green(pigment)}{rgb}{0.0, 0.65, 0.31}

\definecolor{mygray}{gray}{.9}
\usepackage{booktabs}

\def\eg{\emph{e.g.}}

\makeatletter
\let\NAT@parse\undefined
\makeatother
\usepackage[pagebackref=false,breaklinks=true,colorlinks,bookmarks=false]{hyperref}
\hypersetup{colorlinks=true,linkcolor={red},citecolor={hanpurple},urlcolor={magenta}}

\title{\LARGE \bf
$M^2$-Occ: Resilient 3D Semantic Occupancy Prediction for Autonomous Driving with Incomplete Camera Inputs
}

\author{Kaixin Lin$^{1}$, Kunyu Peng$^{2,*}$, Di Wen$^{2}$, Yufan Chen$^{2}$, Ruiping Liu$^{2}$, and Kailun Yang$^{1,*}$%
\thanks{This work was supported in part by the National Natural Science Foundation of China (Grant No. 62473139 and No. 62388101), in part by the Hunan Provincial Research and Development Project (Grant No. 2025QK3019), in part by the State Key Laboratory of Autonomous Intelligent Unmanned Systems (the opening project number ZZKF2025-2-10), and in part by the Deutsche Forschungsgemeinschaft (DFG, German Research Foundation) – SFB-1574 – 471687386.}
\thanks{$^{1}$The authors are with the School of Artificial Intelligence and Robotics and the National Engineering Research Center of Robot Visual Perception and Control Technology, Hunan University, China (email: kailun.yang@hnu.edu.cn).}%
\thanks{$^{2}$The authors are with the Institute for Anthropomatics and Robotics, Karlsruhe Institute of Technology, Germany (email: kunyu.peng@kit.edu).}%
\thanks{*Corresponding authors: Kailun Yang and Kunyu Peng.}%
}

\definecolor{clr_barrier}{RGB}{255, 120, 50}
\definecolor{clr_bicycle}{RGB}{255, 192, 203}
\definecolor{clr_bus}{RGB}{255, 255, 0}
\definecolor{clr_car}{RGB}{0, 150, 245}
\definecolor{clr_const_veh}{RGB}{0, 255, 255}
\definecolor{clr_motorcycle}{RGB}{200, 180, 0}
\definecolor{clr_pedestrian}{RGB}{255, 0, 0}
\definecolor{clr_traffic_cone}{RGB}{255, 240, 150}
\definecolor{clr_trailer}{RGB}{135, 60, 0}
\definecolor{clr_truck}{RGB}{160, 32, 240}
\definecolor{clr_drive_surf}{RGB}{255, 0, 255}
\definecolor{clr_other_flat}{RGB}{139, 137, 137}
\definecolor{clr_sidewalk}{RGB}{75, 0, 75}
\definecolor{clr_terrain}{RGB}{150, 240, 80}
\definecolor{clr_manmade}{RGB}{230, 230, 250}
\definecolor{clr_vegetation}{RGB}{0, 175, 0}

\newcommand{\clsbox}[1]{{\color{#1}\rule{1.2ex}{1.2ex}}~}

\begin{document}

\maketitle
\thispagestyle{empty}
\pagestyle{empty}

\begin{abstract}
Semantic occupancy prediction enables dense 3D geometric and semantic understanding for autonomous driving. However, existing camera-based approaches implicitly assume complete surround-view observations, an assumption that rarely holds in real-world deployment due to occlusion, hardware malfunction, or communication failures. We study semantic occupancy prediction under incomplete multi-camera inputs and introduce $M^2$-Occ, a framework designed to preserve geometric structure and semantic coherence when views are missing. $M^2$-Occ addresses two complementary challenges. First, a Multi-view Masked Reconstruction (MMR) module leverages the spatial overlap among neighboring cameras to recover missing-view representations directly in the feature space. Second, a Feature Memory Module (FMM) introduces a learnable memory bank that stores class-level semantic prototypes. By retrieving and integrating these global priors, the FMM refines ambiguous voxel features, ensuring semantic consistency even when observational evidence is incomplete. We introduce a systematic missing-view evaluation protocol on the nuScenes-based SurroundOcc benchmark, encompassing both deterministic single-view failures and stochastic multi-view dropout scenarios. Under the safety-critical missing back-view setting, $M^2$-Occ improves the IoU by $4.36\%$. As the number of missing cameras increases, the robustness gap further widens; for instance, under the setting with five missing views, our method boosts the IoU by $6.67\%$. These gains are achieved without compromising full-view performance. The source code will be publicly released at \url{https://github.com/qixi7up/M2-Occ}.
\end{abstract}

\section{Introduction}
Autonomous vehicles require a fine-grained understanding of 3D environments for safe navigation decisions~\cite{geiger2012we, caesar2020nuscenes}. 
Recently, 3D semantic occupancy prediction has emerged as a key task, providing a voxel-level representation of free space and semantic obstacles around the ego-vehicle~\cite{cao2022monoscene, wang2023openoccupancy}. 
While LiDAR-based methods provide accurate depth information, camera-based solutions have garnered increasing attention due to dense information and effectiveness~\cite{li2023voxformer, li2022bevformer}.
Semantic Occupancy Prediction extends Bird's-Eye-View (BEV) perception to dense 3D voxel space, thereby distinguishing arbitrarily shaped obstacles and detailed scene semantics, \eg, SurroundOcc~\cite{wei2023surroundocc} and TPVFormer~\cite{huang2023tri}.

\begin{figure}[!t]
    \centering
    \includegraphics[width=0.5\textwidth]{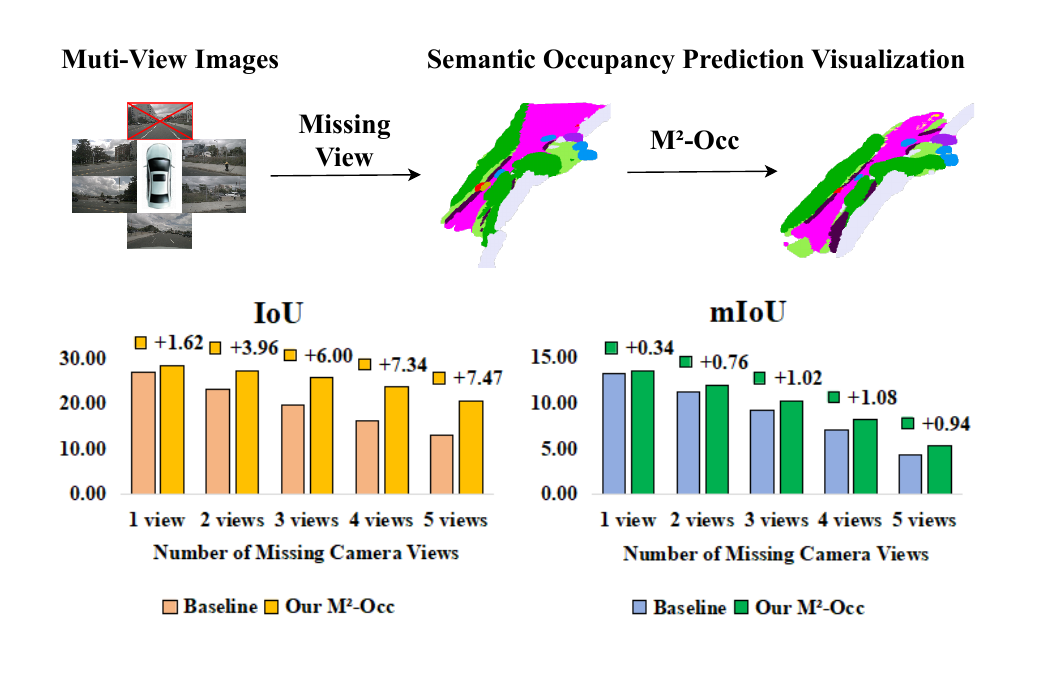}
    \vskip-1ex
    \caption{Existing methods fail with missing camera inputs (\eg, FRONT-view). Our $M^2$-Occ reconstructs missing features from adjacent overlaps and stabilizes semantics via global memory, achieving superior performance.}
    \vskip-1ex
    \label{fig:figure1_comparison}
\end{figure}

Existing multi-camera paradigms for semantic occupancy prediction, such as SurroundOcc~\cite{wei2023surroundocc} and TPVFormer~\cite{huang2023tri}, typically fuse features from multiple views, \eg, six cameras in nuScenes~\cite{caesar2020nuscenes}, to build a unified 3D scene representation. These approaches generally operate under the implicit assumption of ideal sensing conditions, where all cameras are synchronized, calibrated, and fully functional.

In practice, however, real-world scenarios frequently violate this assumption. Camera sensors are inherently vulnerable to complete failure, \eg, lens occlusion, hardware malfunction, or communication dropouts. Such missing-view conditions introduce incomplete spatial coverage and inconsistent cross-view correspondences, which can significantly degrade the stability and reliability of occupancy prediction systems.
In our preliminary experiments, we observed that even representative models like SurroundOcc~\cite{wei2023surroundocc} and TPVFormer~\cite{huang2023tri} suffer a drastic performance drop when a single critical view is lost. This vulnerability poses a serious safety risk for autonomous systems. 
As illustrated in Fig.~\ref{fig:figure1_comparison}, such failures create significant geometric gaps in the perceived environment, compromising the vehicle's situational awareness.
We adopt SurroundOcc~\cite{wei2023surroundocc} and TPVFormer~\cite{huang2023tri} as our primary baselines because they are representative and widely adopted multi-camera occupancy prediction models, achieving strong performance under full-view conditions while being clearly vulnerable to missing inputs---making them ideal testbeds for evaluating robustness enhancements.

To tackle this challenge, we draw inspiration from the remarkable human capability of \emph{amodal spatial perception}: even when a portion of the visual field is occluded or unavailable, humans can effortlessly infer the likely geometry and semantics of the hidden regions by leveraging contextual cues from the surrounding visible areas and prior knowledge accumulated from past experience. This cognitive process combines two complementary mechanisms: (i) \emph{spatial imagination} that extrapolates structural continuity across visual boundaries, and (ii) \emph{semantic memory} that retrieves learned object prototypes to resolve ambiguities in uncertain regions. Motivated by these principles, we propose $M^2$-Occ, a generic framework designed to mitigate the impact of missing sensory information in occupancy prediction. Our approach is built on two pillars: feature-level reconstruction from neighboring views and semantic regularization through a global class-level memory.

First, in typical sensor configurations, adjacent cameras have overlapping Fields of View (FoV). For instance, a front-left camera partially overlaps the front camera's blind spot. 
Leveraging this redundancy, we introduce a Multi-View Masked Reconstruction (MMR) module as a ``soft repair'' mechanism. 
Unlike generative methods~\cite{wang2023drivedreamer} that hallucinate raw pixels, MMR operates in the feature space. It utilizes a transformer-based decoder to aggregate contextual information from neighboring unmasked views to reconstruct the lost features.
Secondly, relying solely on visual reconstruction can yield noisy or ambiguous results, especially when the missing region contains objects with high intra-class variability or when multiple adjacent views are simultaneously lost. 
To provide high-level semantic guidance, we introduce a Feature Memory Module (FMM) that maintains a learnable global memory bank of class-level semantic prototypes. 
The Single-Proto strategy maintains one global centroid per semantic class to capture its core characteristics, promoting stability and robustness under incomplete observations. 
In contrast, the Multi-Proto strategy learns multiple sub-prototypes for each class to model intra-class variance (\textit{e.g.}, different vehicle types or orientations) and dynamically retrieves them based on feature similarity, enabling finer-grained semantic refinement. 
This memory bank serves as a prior knowledge base, allowing the model to refine the reconstructed voxels based on learned class-specific attributes, ensuring that a ``car'' object still retains the characteristic features of a car, even if its visual features are partially corrupted.

To verify the effectiveness of our approach, we provide a systematic set of analyses for missing-view occupancy prediction. We simulate realistic failure patterns, including specific single-view losses (\eg, front or rear camera failure due to physical damage) and stochastic multi-view dropouts. This protocol allows us to quantify the perceptual boundaries of occupancy models and identify the specific vulnerability of existing methods when facing blind spots caused by sensor malfunctions. 
Extensive empirical results on the nuScenes dataset demonstrate that $M^2$-Occ significantly enhances robustness without sacrificing standard performance. Specifically, under the safety-critical ``missing back view'' setting, our method improves the IoU by $4.36\%$ compared to the SurroundOcc~\cite{wei2023surroundocc} baseline, effectively recovering geometry in the rear blind spot. Furthermore, in extreme scenarios where up to $5$ cameras are disabled, our framework exhibits remarkable resilience, maintaining an IoU of $20.24\%$ while the SurroundOcc~\cite{wei2023surroundocc} baseline collapses to $13.57\%$, proving its capability to preserve essential structural information under catastrophic sensor failure. 
Our main contributions are summarized as follows:
\begin{itemize}
\item We systematically study semantic occupancy prediction under incomplete multi-camera inputs, showing that even a single missing view severely degrades state-of-the-art models like SurroundOcc~\cite{wei2023surroundocc} and TPVFormer~\cite{huang2023tri}, motivating robust perception design.

\item We propose $M^2$-Occ with two main components: MMR reconstructs missing-view features from adjacent overlaps, and FMM refines voxel features using learnable semantic prototypes.

\item Extensive experiments on nuScenes demonstrate that our method significantly improves robustness across various missing-view scenarios.
\end{itemize}

\section{Related Work}

\subsection{Semantic Occupancy Prediction}
Semantic Occupancy Prediction (SOP) estimates a dense 3D voxel grid, assigning occupancy and semantic labels to both visible and occluded regions~\cite{song2017semantic, behley2019semantickitti}. This representation enables holistic scene understanding beyond 2D projections or 3D bounding boxes for autonomous driving. Early SOP methods, \eg, LMSCNet~\cite{roldao2020lmscnet}, JS3CNet~\cite{yan2021sparse}, use LiDAR or depth with 3D CNNs to complete sparse scans. LMSCNet~\cite{roldao2020lmscnet} fills occlusions and JS3CNet~\cite{yan2021sparse} introduces context-shape priors, but LiDAR-based approaches face point sparsity and sensor cost. Regarding camera-based SOPs, MonoScene~\cite{cao2022monoscene} shows that a monocular image can reconstruct a scene, while multi-camera $360^{\circ}$ systems like SurroundOcc~\cite{wei2023surroundocc} provide $360^{\circ}$ coverage. Transformer models, \eg, TPVFormer~\cite{huang2023tri}, OccFormer~\cite{Zhang2023OccFormer}, VoxFormer~\cite{li2023voxformer}, COTR~\cite{Ma_2024_CVPR}, introduce 2D–3D fusion techniques further improving accuracy. To further improve performance and robustness, some works fuse multiple sensor modalities so their complementary strengths can compensate for one another. For instance, OpenOccupancy~\cite{Wang_2023_openoccupancy} shows LiDAR-camera fusion outperforms single-sensor, while OccFusion~\cite{ming2024occfusion} fuses images with LiDAR for superior accuracy. Furthermore, FusionOcc~\cite{zhang2024fusionocc} and MS-Occ~\cite{wei2025msocc} show that adding LiDAR or radar improves accuracy but adds complexity.

\begin{figure*}[!t]
    \centering
    \includegraphics[width=1.0\textwidth]{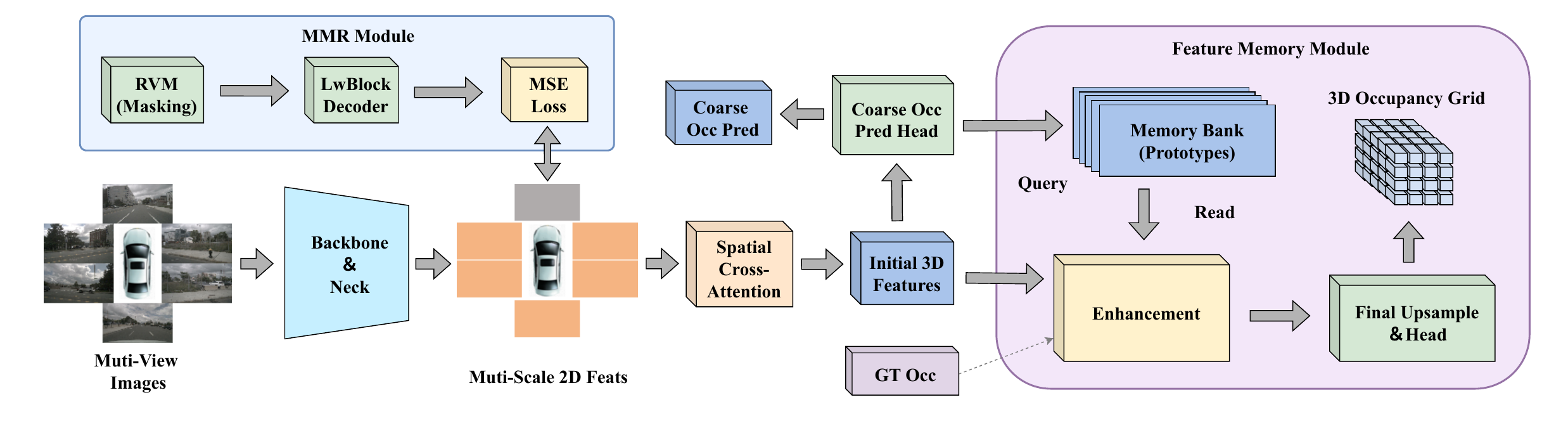}
    \vskip-1ex
    \caption{Overview of $M^2$-Occ. Multi-view images are processed by a shared backbone. MMR reconstructs missing views from adjacent overlaps. Features are lifted to a unified 3D volume. FMM refines voxel features via global semantic prototypes before generating dense 3D occupancy.}
    \vskip-1ex
    \label{fig:pipeline}
\end{figure*}

Meanwhile, semantic occupancy research continues to expand. POP3D~\cite{vobecky2023pop3d} explores open-vocabulary 3D occupancy by aligning visual features with language models, while QuadricFormer~\cite{zuo2025quadricformer} improves efficiency through compact object-centric superquadric representations. Temporal extensions such as UniOcc~\cite{wang2025uniocc} and STCOcc~\cite{liao2025stcocc} incorporate time-series information to predict occupancy dynamics.

Despite these advances, most existing methods assume all camera views are available at inference time. In practice, autonomous systems may encounter missing or occluded inputs. Our work addresses this limitation by explicitly enhancing robustness to sensor failure scenarios.

\subsection{Missing-View BEV Perception}
Robustness to incomplete sensors has been actively studied in BEV-based perception and mapping. UniBEV~\cite{wang2024unibev} is designed to remain functional under missing sensor modalities. MetaBEV~\cite{ge2023metabev} addresses sensor corruptions and modality absence for joint BEV tasks. M-BEV~\cite{chen2024m_bev} is explicitly trained with whole-view masking and reconstructs missing-view features using cross-view context. Beyond detection, SafeMap~\cite{hao2025safemap} strengthens BEV map construction by correcting BEV representations from complete observations, whereas FlexMap~\cite{wang2026flexmap} targets missing views without per-configuration retraining. 

In contrast to these BEV-oriented works, our work focuses on dense 3D semantic occupancy prediction under missing surround-view cameras and introduces feature-level view recovery together with voxel-level semantic regularization to maintain consistent volumetric semantics.

\subsection{Masked Visual Modeling}
Masked visual modeling has emerged as a powerful self-supervised learning paradigm across various domains. Inspired by Masked Autoencoders (MAE)~\cite{he2022masked} in natural language processing and computer vision, recent works~\cite{lin2024bevmae} adapt this framework to BEV-based perception. For instance, MAE-style approaches in BEV, such as BEVT~\cite{wang2022bevt} and VideoMAE~\cite{tong2022videomae}, employ patch-level masking to pre-train models for scalable feature learning. However, these methods primarily focus on general representation learning and do not address the specific challenges of missing or corrupted sensor inputs in autonomous driving.

To address this, the M-BEV framework~\cite{chen2024m_bev} introduces a novel Masked View Reconstruction (MVR) module tailored specifically for BEV perception. Unlike MAE, which masks random patches within a single image, MVR masks the entire camera view and reconstructs it using the spatiotemporal context of neighboring views. This design is crucial for handling sensor failures in the real world, where entire views may be missing. Furthermore, while previous research has explored cross-modal fusion (\textit{e.g.}, MetaBEV~\cite{ge2023metabev}) to mitigate sensor failures, the M-BEV framework achieves robustness using only camera input, making it more cost-effective and scalable.

Masked visual modeling has demonstrated strong potential in BEV perception and related vision tasks, yet it remains underexplored in semantic occupancy prediction. This task is vital for autonomous driving, providing comprehensive 3D scene understanding for safe navigation.

However, robustness under missing-view conditions—especially those caused by sensor failure—has received limited attention. This paper specifically targets this critical scenario to ensure reliable perception under hardware malfunctions or communication dropouts.

\section{Method}

\subsection{Overall Architecture} As shown in Fig.~\ref{fig:pipeline}, the proposed $M^2$-Occ framework is designed to lift multi-view 2D images into a dense 3D semantic occupancy representation while ensuring robustness against sensor failures. 
The architecture follows a standard 2D-to-3D paradigm. 
Let $\mathcal{I} = \{I_i\}_{i=1}^N$ denote the set of input images from $N$ surround-view cameras. First, a shared image backbone with a Feature Pyramid Network serves as the encoder $\mathcal{E}$ to extract multi-scale 2D features $F_{2D} = \mathcal{E}(\mathcal{I})$. 
Subsequently, a 2D-to-3D view transformation module $\mathcal{T}$, based on spatial cross-attention, lifts these features into a unified 3D volume space, yielding $V_{3D} = \mathcal{T}(F_{2D})$. 
Finally, a 3D occupancy head $\mathcal{H}$ processes the volume to predict the voxel-wise semantic labels $Y$. The entire pipeline can be formulated as:
\begin{equation}
Y = \mathcal{H}(V_{3D}) = \mathcal{H}(\mathcal{T}(\mathcal{E}(\mathcal{I})))
\end{equation}
To mitigate the impact of missing views, we introduce the Multi-view Masked Reconstruction (MMR) module during the feature extraction stage $\mathcal{E}(\cdot)$ and the Feature Memory Module (FMM) during the volume refinement stage $\mathcal{T}(\cdot)$.

\begin{figure}[!t]
    \centering
    \includegraphics[width=0.5\textwidth]{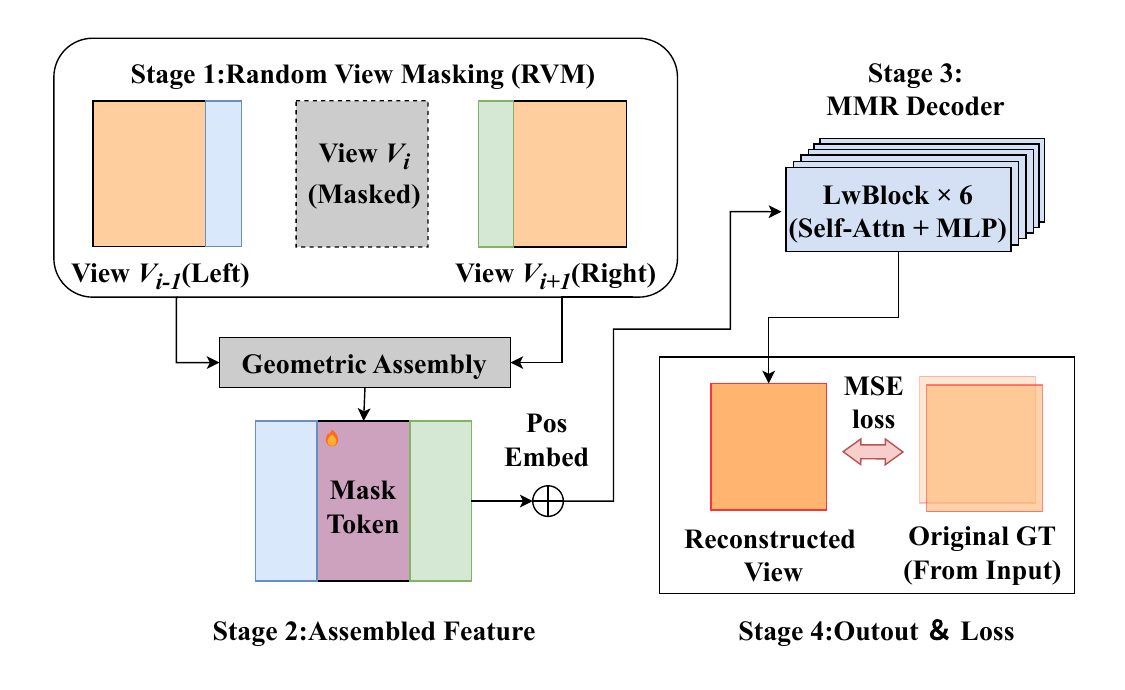}
    \vskip-1ex
    \caption{An overview of Multi-view Masked Reconstruction (MMR). The MMR module extracts overlapping boundary features from adjacent unmasked views and concatenates them with a central learnable mask token. A lightweight transformer decoder then processes this structural prior to reconstruct the missing view's representations, preserving spatial continuity. 
    }
    \vskip-1ex
    \label{fig:reconstruction}
\end{figure}

\subsection{Multi-view Masked Reconstruction (MMR)}
The structure of the Multi-view Masked Reconstruction (MMR) module is shown in Fig.~\ref{fig:reconstruction}. 

\subsubsection{Perspective Relationship Modeling}
In surround-view driving setups, cameras are typically mounted with significant overlaps to cover the full $360^{\circ}$ field of view. We explicitly model this physical layout as a cyclic graph to exploit spatial redundancy. For any camera view $v_i$, we identify its spatial neighborhood $\mathcal{N}(v_i)$ consisting of the immediately adjacent left and right cameras:
\begin{equation}
\mathcal{N}(v_i) = \{ v_{(i-1)\pmod N}, \ v_{(i+1) \pmod N} \}.
\end{equation}
This graph allows us to locate the specific sources of complementary visual information when a particular view suffers from occlusion or failure.
\subsubsection{Overlap-based Feature Aggregation}
When a specific view $v_i$ is masked (simulating a failure), we cannot estimate the environmental information solely based on its own visual input. However, the boundaries of the missing view are often visible in the neighboring cameras. To leverage this contextual information, we perform a feature cropping and splicing operation. Let $\mathbf{f}_{left}$ and $\mathbf{f}_{right}$ represent the feature maps of the left and right neighbors, respectively. We extract the overlapping boundary regions of width $w_{ov}$ (which corresponds to the physical overlap area) and concatenate them with a learnable mask token $\mathbf{e}_{mask}$. This synthesized feature $\mathbf{f}_{ref}$ serves as a structural prior for reconstruction:
\begin{equation}
\mathbf{f}_{ref} = \text{Concat}( \mathbf{f}_{left}[:, -w_{ov}:], \ \mathbf{e}_{mask}, \ \mathbf{f}_{right}[:, :w_{ov}] ), 
\end{equation}
where $\text{Concat}(\cdot)$ denotes channel-wise concatenation of the feature slices. Here, the mask token acts as a placeholder for the central blind spot, initializing the query for the subsequent generative process.

\begin{figure}[!t]
    \centering
    \includegraphics[width=0.5\textwidth]{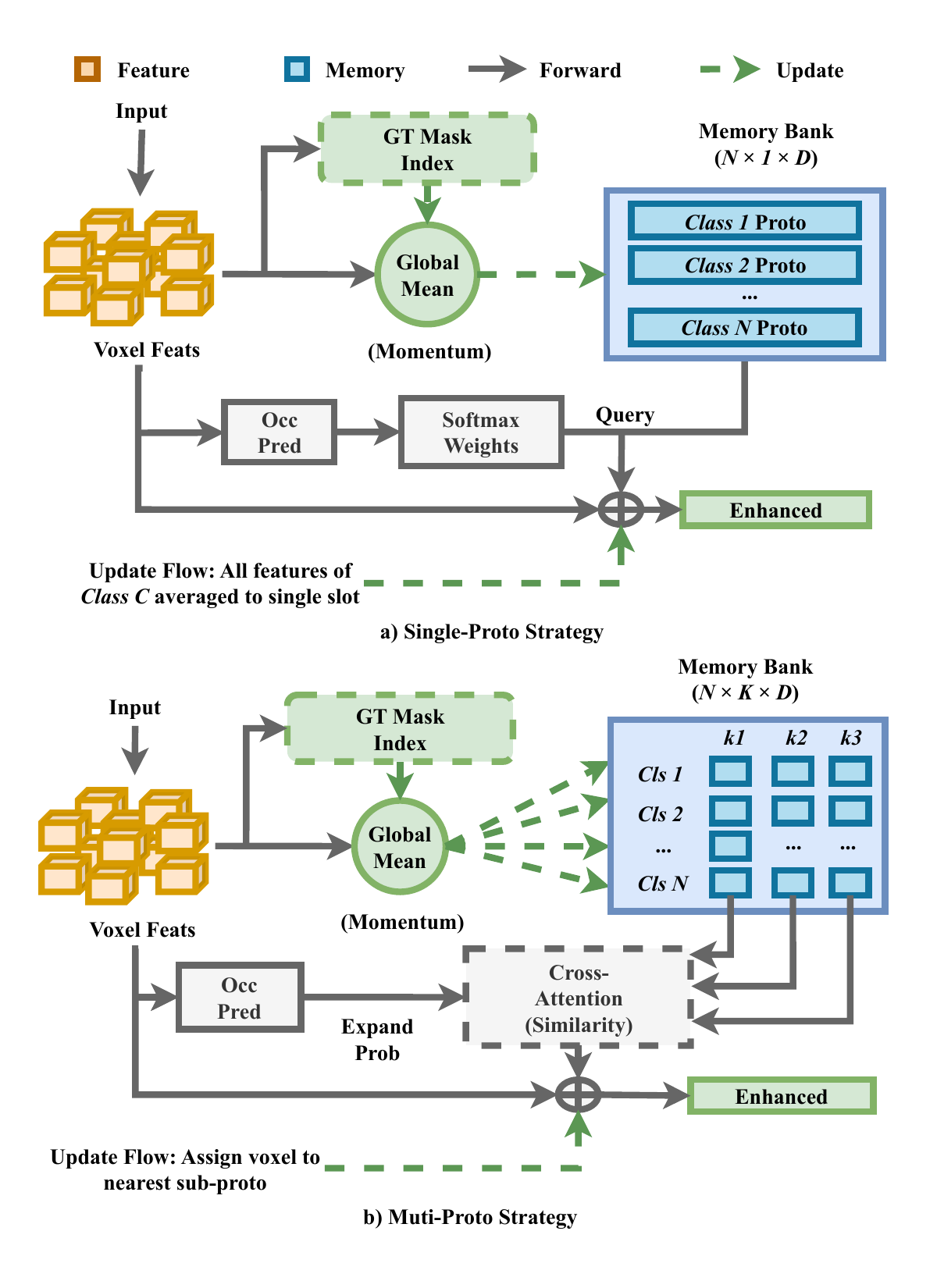}
    \vskip-1ex
    \caption{A comparison between the single-proto strategy and the multi-proto strategy. While the single-proto approach maintains one global centroid per semantic class, the multi-proto strategy captures intra-class variance by learning multiple sub-prototypes and dynamically retrieving them based on feature similarity.}
    \label{fig:proto_variant}
    \vskip-1ex
\end{figure}

\subsubsection{Masked Reconstruction Mechanism}
To reconstruct missing details from the rough structural prior, we employ a lightweight transformer decoder $\mathcal{D}$. 
Since geometric correspondence is crucial, we augment the reference features $\mathbf{f}_{ref}$ with a learnable positional embedding $\mathbf{p}_{pos}$ to preserve spatial awareness. 
The decoder, composed of stacked residual transformer layers with layer scaling, refines the features to approximate the original unmasked representation
$\hat{\mathbf{f}}_i$:
\begin{equation}
\hat{\mathbf{f}}_i = \mathcal{D}( \mathbf{f}_{ref} + \mathbf{p}_{pos} ).
\end{equation}
This process forces the network to learn the spatial continuity of the environment, enabling it to infer the contents of a missing view based on context.

\subsubsection{MMR Loss Function}
To ensure the reconstructed features are semantically meaningful and aligned with the original feature distribution, we impose a reconstruction constraint between the estimated unmasked representation $\hat{\mathbf{f}}_i$ and the original unmasked representation $\mathbf{f}^{gt}_i$. Crucially, we calculate the loss only on the set of masked view indices $\mathcal{M}$ to prevent the network from learning identity mappings for unmasked views. We use the Mean Squared Error (MSE) loss:
\begin{equation}
\mathcal{L}_{MMR} = \frac{1}{|\mathcal{M}|} \sum_{i \in \mathcal{M}} \left\lVert \hat{\mathbf{f}}_i - \mathbf{f}_i^{\mathrm{gt}} \right\rVert_2^2.
\end{equation}

\subsection{Feature Memory Module (FMM)}

\begin{table*}[!t]
\centering
\caption{Single-view missing results on SurroundOcc (IoU, mIoU, as well as per-category voxel-level IoU).}
\label{tab:main_results}

\resizebox{\textwidth}{!}{%
\begin{tabular}{l|l|cc|*{16}{c}}
\toprule
\textbf{Setting} & \textbf{Method} & \textbf{IoU} & \textbf{mIoU} &
\rotatebox{90}{\clsbox{clr_barrier}barrier} & 
\rotatebox{90}{\clsbox{clr_bicycle}bicycle} & 
\rotatebox{90}{\clsbox{clr_bus}bus} & 
\rotatebox{90}{\clsbox{clr_car}car} &
\rotatebox{90}{\clsbox{clr_const_veh}const.veh.} & 
\rotatebox{90}{\clsbox{clr_motorcycle}motorcycle} & 
\rotatebox{90}{\clsbox{clr_pedestrian}pedestrian} & 
\rotatebox{90}{\clsbox{clr_traffic_cone}traffic cone} &
\rotatebox{90}{\clsbox{clr_trailer}trailer} & 
\rotatebox{90}{\clsbox{clr_truck}truck} & 
\rotatebox{90}{\clsbox{clr_drive_surf}drive.surf.} & 
\rotatebox{90}{\clsbox{clr_other_flat}other flat} &
\rotatebox{90}{\clsbox{clr_sidewalk}sidewalk} & 
\rotatebox{90}{\clsbox{clr_terrain}terrain} & 
\rotatebox{90}{\clsbox{clr_manmade}manmade} & 
\rotatebox{90}{\clsbox{clr_vegetation}vegetation} \\
\midrule

\multirow{2}{*}{Standard}
 & SurroundOcc (Wei~\textit{et al.}~2023~\cite{wei2023surroundocc}) & \textbf{31.18} & 17.23 & 16.16 & 8.15 & 19.07 & 29.28 & 10.22 & 9.11 & 10.59 & 8.19 & 11.09 & 18.54 & 38.04 & 19.20 & 23.35 & 21.57 & 12.68 & 20.44 \\
 & Ours (SurroundOcc) & 31.17 & \textbf{18.03} & 17.56 & 7.65 & 23.86 & 29.54 & 9.92 & 12.63 & 11.31 & 7.96 & 12.03 & 20.31 & 38.78 & 18.29 & 24.10 & 21.42 & 12.63 & 20.56 \\
\midrule

\multirow{2}{*}{Front}
 & SurroundOcc (Wei~\textit{et al.}~2023~\cite{wei2023surroundocc}) & 23.96 & 13.35 & 13.90 & 7.83 & 15.91 & 25.02 & 9.90 & 6.01 & 8.93 & 8.04 & 9.55 & 16.77 & 14.56 & 14.12 & 16.41 & 17.39 & 11.34 & 17.96 \\
 & Ours (SurroundOcc) & \textbf{29.30} & \textbf{15.19} & 14.64 & 4.03 & 18.64 & 25.26 & 9.40 & 7.48 & 7.41 & 6.92 & 10.27 & 17.56 & 35.35 & 14.09 & 21.57 & 19.24 & 11.91 & 19.33 \\
\midrule

\multirow{2}{*}{Front Right}
 & SurroundOcc (Wei~\textit{et al.}~2023~\cite{wei2023surroundocc}) & 29.89 & 15.92 & 15.12 & 7.28 & 17.66 & 27.23 & 8.65 & 8.21 & 9.40 & 7.49 & 10.29 & 17.43 & 36.58 & 17.82 & 21.45 & 19.29 & 11.80 & 19.01 \\
 & Ours (SurroundOcc) & \textbf{30.34} & \textbf{17.02} & 16.32 & 6.99 & 23.64 & 27.58 & 8.71 & 11.75 & 9.83 & 7.38 & 11.03 & 19.26 & 37.79 & 17.31 & 23.03 & 20.38 & 11.82 & 19.54 \\
\midrule

\multirow{2}{*}{Front Left}
 & SurroundOcc (Wei~\textit{et al.}~2023~\cite{wei2023surroundocc}) & 29.83 & 15.95 & 15.41 & 8.02 & 17.91 & 27.60 & 7.78 & 8.83 & 9.32 & 7.32 & 10.49 & 16.72 & 36.69 & 17.51 & 21.51 & 19.72 & 11.78 & 18.67 \\
 & Ours (SurroundOcc) & \textbf{30.49} & \textbf{16.96} & 16.70 & 6.42 & 23.02 & 28.09 & 8.23 & 11.50 & 9.62 & 7.37 & 11.26 & 18.63 & 37.96 & 17.37 & 23.02 & 20.54 & 11.98 & 19.71 \\
\midrule

\multirow{2}{*}{Back}
 & SurroundOcc (Wei~\textit{et al.}~2023~\cite{wei2023surroundocc}) & 23.51 & 13.42 & 12.99 & 10.03 & 14.44 & 22.40 & 9.21 & 8.82 & 8.68 & 7.40 & 10.23 & 15.63 & 28.11 & 13.39 & 18.42 & 13.98 & 10.28 & 10.64 \\
 & Ours (SurroundOcc) & \textbf{27.87} & \textbf{14.20} & 14.24 & 2.91 & 17.22 & 22.40 & 9.06 & 8.41 & 5.96 & 5.47 & 10.73 & 16.21 & 34.24 & 13.56 & 20.66 & 18.06 & 11.14 & 16.91 \\
\midrule

\multirow{2}{*}{Back Left}
 & SurroundOcc (Wei~\textit{et al.}~2023~\cite{wei2023surroundocc}) & 29.74 & 15.99 & 15.51 & 6.41 & 18.46 & 27.35 & 9.24 & 8.12 & 9.60 & 7.92 & 10.34 & 17.30 & 36.59 & 17.27 & 21.36 & 19.83 & 11.60 & 18.97 \\
 & Ours (SurroundOcc) & \textbf{30.29} & \textbf{17.19} & 17.19 & 6.74 & 23.79 & 27.86 & 9.53 & 11.73 & 10.06 & 7.52 & 11.28 & 19.20 & 38.04 & 17.80 & 23.04 & 20.31 & 11.71 & 19.18 \\
\midrule

\multirow{2}{*}{Back Right}
 & SurroundOcc (Wei~\textit{et al.}~2023~\cite{wei2023surroundocc}) & 30.06 & 16.08 & 14.32 & 7.12 & 17.67 & 27.73 & 9.01 & 8.55 & 9.81 & 7.49 & 10.91 & 17.47 & 36.90 & 18.45 & 21.74 & 19.49 & 11.74 & 18.80 \\
 & Ours (SurroundOcc) & \textbf{30.38} & \textbf{17.17} & 16.19 & 6.69 & 23.48 & 28.23 & 8.75 & 11.95 & 10.48 & 7.46 & 11.76 & 19.20 & 38.18 & 17.71 & 23.28 & 20.36 & 11.75 & 19.31 \\
\bottomrule
\end{tabular}
}
\vskip-1ex
\end{table*}

While MMR recovers the geometric structure, the reconstructed features may still suffer from blurring or semantic ambiguity. The FMM addresses this by introducing a global memory bank $\mathbf{M}$ that stores high-quality semantic prototypes, acting as a ``long-term memory'' to refine the transient observations. A comparison between the Single-Proto and Multi-Proto strategies is provided in Figure~\ref{fig:proto_variant}. 
\subsubsection{Single-Proto Strategy}
In this simplified strategy, we assume each semantic class $k$ can be represented by a single global centroid. 
We maintain a prototype $\mathbf{m}_k$ which acts as the ideal feature representation for class $k$. 
To ensure stability during training, we update the prototype using a momentum moving average of the mean feature $\bar{\mathbf{f}}_k$ of all voxels assigned to class $k$ in the current batch:
\begin{equation}
\mathbf{m}_k^{(t)} = (1-\lambda) \mathbf{m}_k^{(t-1)} + \lambda \cdot \bar{\mathbf{f}}_k,
\end{equation}
where $\lambda$ is a momentum coefficient set to 0.1. 
This effectively filters out noise and outliers from individual mini-batches.
\subsubsection{Multi-Proto Strategy}
Real-world objects exhibit high intra-class variance (\textit{e.g.}, a ``truck'' can be a pickup or a semi-trailer). 
To capture this diversity, we extend the memory to store $N_p$ sub-prototypes per class.

Similarity Calculation: For a query voxel feature $\mathbf{x}$, we compute its cosine similarity with all sub-prototypes $\mathbf{m}_{k,j}$ to determine which subclass it likely belongs to:
\begin{equation}
s_{k,j} = \frac{\mathbf{x} \cdot \mathbf{m}_{k,j}}{| \mathbf{x} | | \mathbf{m}_{k,j} |}.
\end{equation}
Softmax Weighting: These similarity scores are normalized via a softmax function with temperature $\tau$ to produce retrieval weights $\alpha_{k,j}$, where a lower temperature emphasizes the most relevant prototypes while higher values yield a smoother distribution:
\begin{equation}
\alpha_{k,j} = \frac{\exp(s_{k,j} / \tau)}{\sum_{j'} \exp(s_{k,j'} / \tau)}.
\end{equation}

\subsubsection{Memory-Enhanced Feature}
Finally, we inject the retrieved semantic knowledge back into the 3D volume. Using the predicted class probability $P(k)$ as a gate, we aggregate the weighted prototypes and add them to the original feature $\mathbf{x}$ as a residual correction:
\begin{equation}
\mathbf{x}' = \mathbf{x} + \sum_{k=1}^K \left(P(k) \sum_{j=1}^{N_p} \alpha_{k,j} \mathbf{m}_{k,j}\right),
\end{equation}
where $K$ is the number of semantic classes, $P(k)$ denotes the predicted probability for class $k$, and $\alpha_{k,j}$ are the retrieval weights for sub-prototypes. This step significantly sharpens the semantic boundaries, especially in regions reconstructed by MMR, where the original features might be noisy.

\section{Experiments}
In this section, we conduct comprehensive experiments to evaluate the proposed $M^2$-Occ. 
Table~\ref{tab:main_results} presents the comparison under various single-view missing scenarios. Table~\ref{tab:tpvformer_results} presents the TPVFormer results with both voxel-level and point-level mIoU. Table~\ref{tab:ablation_num} analyzes the robustness as the number of missing cameras increases. Table~\ref{tab:ablation_module} ablates the contribution of each module. Table~\ref{tab:compute_overhead} reports the computational overhead.

\begin{table*}[!t]
\centering
\caption{Single-view missing results on TPVFormer (voxel-level and point-level mIoU, as well as per-category IoU).}
\label{tab:tpvformer_results}

\resizebox{\textwidth}{!}{%
\begin{tabular}{l|l|c|c|*{16}{c}}
\toprule
\textbf{Setting} & \textbf{Method} & \textbf{mIoU (vox)} & \textbf{mIoU (pts)} &
\rotatebox{90}{\clsbox{clr_barrier}barrier} & 
\rotatebox{90}{\clsbox{clr_bicycle}bicycle} & 
\rotatebox{90}{\clsbox{clr_bus}bus} & 
\rotatebox{90}{\clsbox{clr_car}car} &
\rotatebox{90}{\clsbox{clr_const_veh}const.veh.} & 
\rotatebox{90}{\clsbox{clr_motorcycle}motorcycle} & 
\rotatebox{90}{\clsbox{clr_pedestrian}pedestrian} & 
\rotatebox{90}{\clsbox{clr_traffic_cone}traffic cone} &
\rotatebox{90}{\clsbox{clr_trailer}trailer} & 
\rotatebox{90}{\clsbox{clr_truck}truck} & 
\rotatebox{90}{\clsbox{clr_drive_surf}drive.surf.} & 
\rotatebox{90}{\clsbox{clr_other_flat}other flat} &
\rotatebox{90}{\clsbox{clr_sidewalk}sidewalk} & 
\rotatebox{90}{\clsbox{clr_terrain}terrain} & 
\rotatebox{90}{\clsbox{clr_manmade}manmade} & 
\rotatebox{90}{\clsbox{clr_vegetation}vegetation} \\
\midrule

\multirow{2}{*}{Standard}
 & TPVFormer (Huang~\textit{et al.}~2023~\cite{huang2023tri}) & 47.84 & 26.92 & 35.07 & 12.71 & 48.70 & 48.51 & 29.52 & 20.04 & 19.61 & 8.90 & 38.32 & 46.09 & 15.45 & 13.82 & 19.82 & 20.83 & 27.29 & 25.99 \\
 & Ours (TPVFormer) & \textbf{50.26} & \textbf{30.11} & 36.42 & 11.30 & 50.71 & 53.86 & 35.98 & 32.90 & 28.90 & 10.62 & 41.51 & 53.28 & 17.41 & 12.75 & 19.18 & 21.45 & 28.49 & 27.06 \\
\midrule

\multirow{2}{*}{Front}
 & TPVFormer (Huang~\textit{et al.}~2023~\cite{huang2023tri}) & 43.82 & 25.45 & 34.17 & 12.04 & 39.19 & 47.20 & 29.71 & 17.60 & 17.41 & 8.50 & 32.28 & 44.52 & 18.04 & 13.42 & 19.66 & 21.73 & 26.94 & 24.80 \\
 & Ours (TPVFormer) & \textbf{46.89} & \textbf{27.94} & 34.43 & 10.92 & 40.95 & 52.07 & 34.78 & 32.18 & 27.38 & 10.33 & 36.86 & 48.97 & 16.44 & 11.32 & 17.39 & 20.69 & 27.41 & 24.98 \\
\midrule

\multirow{2}{*}{Front Right}
 & TPVFormer (Huang~\textit{et al.}~2023~\cite{huang2023tri}) & 42.00 & 23.48 & 32.16 & 11.96 & 43.43 & 45.64 & 20.79 & 17.72 & 13.84 & 8.36 & 33.24 & 40.37 & 14.72 & 11.28 & 16.55 & 19.07 & 23.48 & 23.06 \\
 & Ours (TPVFormer) & \textbf{44.36} & \textbf{26.09} & 31.40 & 10.33 & 45.91 & 49.06 & 26.16 & 23.61 & 26.88 & 9.35 & 35.34 & 45.71 & 16.48 & 10.25 & 17.21 & 20.45 & 25.12 & 24.15 \\
\midrule

\multirow{2}{*}{Front Left}
 & TPVFormer (Huang~\textit{et al.}~2023~\cite{huang2023tri}) & 42.96 & 23.62 & 31.51 & 11.83 & 45.36 & 43.90 & 23.69 & 18.04 & 13.67 & 8.36 & 32.99 & 40.87 & 13.52 & 11.72 & 18.86 & 18.75 & 23.41 & 21.51 \\
 & Ours (TPVFormer) & \textbf{45.87} & \textbf{26.91} & 33.08 & 10.85 & 44.40 & 49.34 & 30.00 & 29.11 & 25.54 & 9.66 & 36.66 & 47.58 & 16.96 & 10.34 & 18.03 & 20.70 & 25.36 & 23.00 \\
\midrule

\multirow{2}{*}{Back}
 & TPVFormer (Huang~\textit{et al.}~2023~\cite{huang2023tri}) & 40.66 & 22.36 & 31.43 & 12.34 & 34.52 & 37.01 & 26.32 & 16.30 & 17.09 & 9.00 & 33.88 & 30.38 & 14.49 & 12.13 & 17.69 & 20.27 & 23.14 & 21.77 \\
 & Ours (TPVFormer) & \textbf{43.48} & \textbf{25.41} & 31.64 & 11.12 & 34.37 & 45.16 & 32.01 & 28.23 & 26.70 & 10.08 & 35.73 & 38.92 & 16.71 & 13.00 & 17.38 & 20.11 & 23.57 & 21.86 \\
\midrule

\multirow{2}{*}{Back Left}
 & TPVFormer (Huang~\textit{et al.}~2023~\cite{huang2023tri}) & 41.76 & 22.52 & 30.08 & 11.77 & 45.31 & 35.18 & 25.86 & 17.54 & 7.17 & 8.55 & 30.73 & 40.01 & 18.15 & 12.72 & 19.70 & 18.25 & 21.19 & 18.11 \\
 & Ours (TPVFormer) & \textbf{45.41} & \textbf{26.16} & 31.38 & 11.21 & 45.13 & 43.23 & 32.42 & 28.64 & 23.07 & 9.51 & 35.78 & 47.29 & 15.82 & 11.95 & 17.17 & 20.94 & 24.41 & 20.66 \\
\midrule

\multirow{2}{*}{Back Right}
 & TPVFormer (Huang~\textit{et al.}~2023~\cite{huang2023tri}) & 41.66 & 23.35 & 29.53 & 11.77 & 41.56 & 38.45 & 25.29 & 16.10 & 12.51 & 7.87 & 38.20 & 36.91 & 18.98 & 14.42 & 19.42 & 19.30 & 21.35 & 21.86 \\
 & Ours (TPVFormer) & \textbf{44.98} & \textbf{26.28} & 29.55 & 10.74 & 45.60 & 43.99 & 29.10 & 29.05 & 27.61 & 8.09 & 39.07 & 43.70 & 15.99 & 12.69 & 17.73 & 20.55 & 23.76 & 23.18 \\
\bottomrule
\end{tabular}
}
\end{table*}

\subsection{Datasets and Metrics}
To rigorously evaluate the effectiveness of the proposed $M^2$-Occ framework in both standard perception and various sensor-failure scenarios, we conduct extensive experiments on the nuScenes dataset~\cite{caesar2020nuscenes}, which is a large-scale, multimodal dataset that has become a cornerstone for benchmarking autonomous driving algorithms in complex urban environments. 
Developed by Motional, it is the first dataset to provide full 360-degree sensor suite coverage, comprising data from 6 cameras, 1 LiDAR, 5 RADAR units, and a high-precision GPS/IMU system. 
The dataset contains $1,000$ driving scenes captured in Boston and Singapore, totaling approximately $1.4$ million images and $390k$ LiDAR sweeps. 
With rigorous annotations for $23$ object categories across diverse weather conditions and times of day, nuScenes serves as a primary benchmark for tasks such as 3D object detection, multi-object tracking, and trajectory prediction.

In addition, we adopt the dense voxel annotations generated by SurroundOcc through multi-frame LiDAR point cloud aggregation and Poisson surface reconstruction, instead of sparse 3D bounding boxes, enabling the model to perceive the complete 3D structure and occupancy state of the surrounding environment rather than performing only simple object detection. 

We report the standard metrics for semantic occupancy: Intersection over Union (IoU) for geometric completeness and mean IoU (mIoU) for semantic accuracy. For TPVFormer, we report both voxel-level mIoU and point-level mIoU to provide a comprehensive evaluation.

\subsection{Implementation Details}
\textbf{Implementation Details.} For the SurroundOcc baseline, we follow its official implementation, using ResNet-101~\cite{he2016deep} initialized with FCOS3D~\cite{fcos3d} pre-trained weights as the visual encoder, with a voxel volume of $200 \times 200 \times 16$ covering $[-50m, 50m]$ in the ground plane. For the TPVFormer baseline, we adopt settings consistent with the original paper~\cite{huang2023tri}, using ResNet-101-DCN~\cite{he2016deep} as the visual encoder, with TPV plane resolution of $200 \times 200 \times 16$ and feature dimension of $128$. Both models are trained for $6$ epochs using the AdamW optimizer~\cite{loshchilov2017decoupled} with a learning rate of $2 \times 10^{-4}$ and weight decay of $0.01$, where TPVFormer additionally employs a cosine learning rate scheduler. The lightweight transformer decoder $\mathcal{D}$ in MMR consists of $6$ Transformer blocks, each with $8$ attention heads and an MLP ratio of $4$. To simulate sensor failures, we randomly discard images of corresponding views using our Random View Masking (RVM) module during training, and mask specific views during testing to evaluate robustness. This strategy is applied to both baselines.

\subsection{Robustness to Single-View Failures}

\textbf{SurroundOcc Results.} Table~\ref{tab:main_results} reports results when each individual camera view is removed. Across all single-view failure scenarios, $M^2$-Occ consistently improves geometric IoU compared to the baseline model. Under the safety-critical missing rear-view condition, IoU increases from $23.51\%$ to $27.87\%$ ($+4.36\%$). Similar improvements are observed when removing the front view ($23.96\% \rightarrow 29.30\%$) and front-left view ($29.83\% \rightarrow 30.49\%$).

Per-category analysis shows that $M^2$-Occ achieves substantial gains on large-scale structures (\eg, ``drive.surf.'' improves from $28.11\%$ to $34.24\%$ under missing back-view), while performance on small objects exhibits mixed results (\eg, ``pedestrian'' IoU drops from $8.68\%$ to $5.96\%$, and ``traffic cone'' from $7.40\%$ to $5.47\%$). This degradation stems from two factors: (i) MMR relies on overlapping boundary regions that may lack sufficient resolution for distant or small instances, and (ii) feature-level reconstruction inevitably loses high-frequency details.

\textbf{TPVFormer Results.} Table~\ref{tab:tpvformer_results} reports the performance of the TPVFormer baseline under the same single-view missing scenarios, reporting both voxel-level and point-level mIoU. Consistent with observations on SurroundOcc, $M^2$-Occ consistently improves TPVFormer performance across all missing-view settings. Under the standard full-view setting, our method improves voxel-level mIoU from $47.84\%$ to $50.26\%$ ($+2.42\%$) and point-level mIoU from $26.92\%$ to $30.11\%$ ($+3.19\%$). Under the safety-critical missing back-view setting, voxel-level mIoU improves from $40.66\%$ to $43.48\%$ ($+2.82\%$) and point-level mIoU from $22.36\%$ to $25.41\%$ ($+3.05\%$). Per-category analysis shows that the improvements on TPVFormer are more balanced across categories, with particularly notable gains on small objects such as ``motorcycle'' (point IoU from $20.04\%$ to $32.90\%$) and ``pedestrian'' (point IoU from $19.61\%$ to $28.90\%$). This difference may stem from the fact that the feature-level reconstruction of MMR is more compatible with TPVFormer's tri-perspective view (TPV) representation and cross-attention mechanism. These results further validate the generalization capability and effectiveness of $M^2$-Occ across different architectures.

\subsection{Scaling Behavior under Multi-View Dropout}

\begin{table}[!t]
\centering
\caption{Robustness analysis under different numbers of missing views. We compare the baseline, our method with MMR only, and our full model with both MMR and FMM.}
\label{tab:ablation_num}
\resizebox{\columnwidth}{!}{%
\begin{tabular}{c|c|c|c}
\toprule
\textbf{Missing Views} & \textbf{Method} & \textbf{IoU (\%)} & \textbf{mIoU (\%)} \\
\midrule
\multirow{1}{*}{0} & Baseline & 31.18 & 17.23 \\
\midrule
\multirow{3}{*}{1} & Baseline & 27.70 & 14.95 \\
 & + MMR & 29.70 & 15.75 \\
 & + MMR + FMM & \textbf{29.77} & \textbf{16.25} \\
\midrule
\multirow{3}{*}{3} & Baseline & 20.00 & 10.04 \\
 & + MMR & 26.09 & 11.11 \\
 & + MMR + FMM & \textbf{26.20} & \textbf{11.69} \\
\midrule
\multirow{3}{*}{5} & Baseline & 13.57 & 4.49 \\
 & + MMR & 19.75 & 4.96 \\
 & + MMR + FMM & \textbf{20.24} & \textbf{5.53} \\
\bottomrule
\end{tabular}%
}
\vskip-1ex
\end{table}

We further evaluate performance when multiple cameras are simultaneously removed (Table~\ref{tab:ablation_num}). 
When the baseline model suffers from single-view missing, its IoU and mIoU drop to $27.70\%$ and $14.95\%$, respectively. 
In contrast, our $M^2$-Occ effectively alleviates such performance degradation, achieving $29.77\%$ and $16.25\%$, respectively.

As the number of missing views increases, the robustness gap widens substantially. 
With three missing views, the baseline decreases to $20.00\%$ IoU and $10.04\%$ mIoU, while $M^2$-Occ maintains $26.20\%$ and $11.69\%$. 
Under the most extreme setting of five missing views, the baseline further drops to $13.57\%$ and $4.49\%$, whereas $M^2$-Occ retains $20.24\%$ and $5.53\%$.

Notably, in severe multi-view dropout scenarios, improvements are observed in both IoU and mIoU. 
This indicates that when structural evidence becomes highly sparse, the combination of reconstruction and semantic regularization stabilizes both geometry and core semantic predictions.

\subsection{Ablation Study of the Model Components}

\begin{table}[!t]
\centering
\caption{Ablation study on the modules in the presented framework.}
\label{tab:ablation_module}
\resizebox{\columnwidth}{!}{%
\begin{tabular}{c|ccc|cc}
\toprule
Missing & MMR & FMM & IoU & mIoU \\
\midrule
$\times$ & $\times$ & $\times$ & 31.18 & 17.23 \\
\checkmark & $\times$ & $\times$ & 27.70 & 14.95 \\
\checkmark & \checkmark & $\times$ & 29.70 & 15.75 \\
\checkmark & \checkmark & \checkmark & \textbf{29.77} & \textbf{16.25} \\
\bottomrule
\end{tabular}%
}
\vskip-1ex
\end{table}

Table~\ref{tab:ablation_module} provides a detailed ablation study of the Multi-view Masked Reconstruction (MMR) and Feature Memory Module (FMM) under missing-view conditions.

When a view is masked without any recovery mechanism, the baseline IoU drops from $31.18\%$ to $27.70\%$, indicating that the performance degradation primarily stems from incomplete spatial coverage and broken cross-view correspondences. 
Introducing MMR alone improves IoU to $29.70\%$, recovering $+2.00\%$ over the degraded baseline. This gain confirms that feature-level reconstruction effectively restores structural continuity by exploiting spatial overlap between adjacent cameras. Notably, the improvement is more pronounced in geometric IoU than mIoU, suggesting that MMR mainly compensates for large-scale spatial structures (\eg, road surfaces and vehicle volumes), which dominate occupancy estimation.

The FMM injects global class-level priors to suppress ambiguity in uncertain voxels, especially in regions where the reconstructed features remain noisy or partially incomplete. Incorporating FMM with the single-prototype strategy further increases IoU to $29.77\%$ and mIoU to $16.25\%$, demonstrating the complementary benefits of combining structural reconstruction with semantic regularization.

Overall, the ablation results demonstrate a clear division of labor between the two modules: MMR restores geometric completeness by reconstructing structural features, while FMM enhances feature representation by imposing global class-level constraints. Their combination achieves stable performance under incomplete multi-view observations.

\subsection{Qualitative Results}

\begin{figure}[!t]
    \centering
    \includegraphics[width=0.5\textwidth]{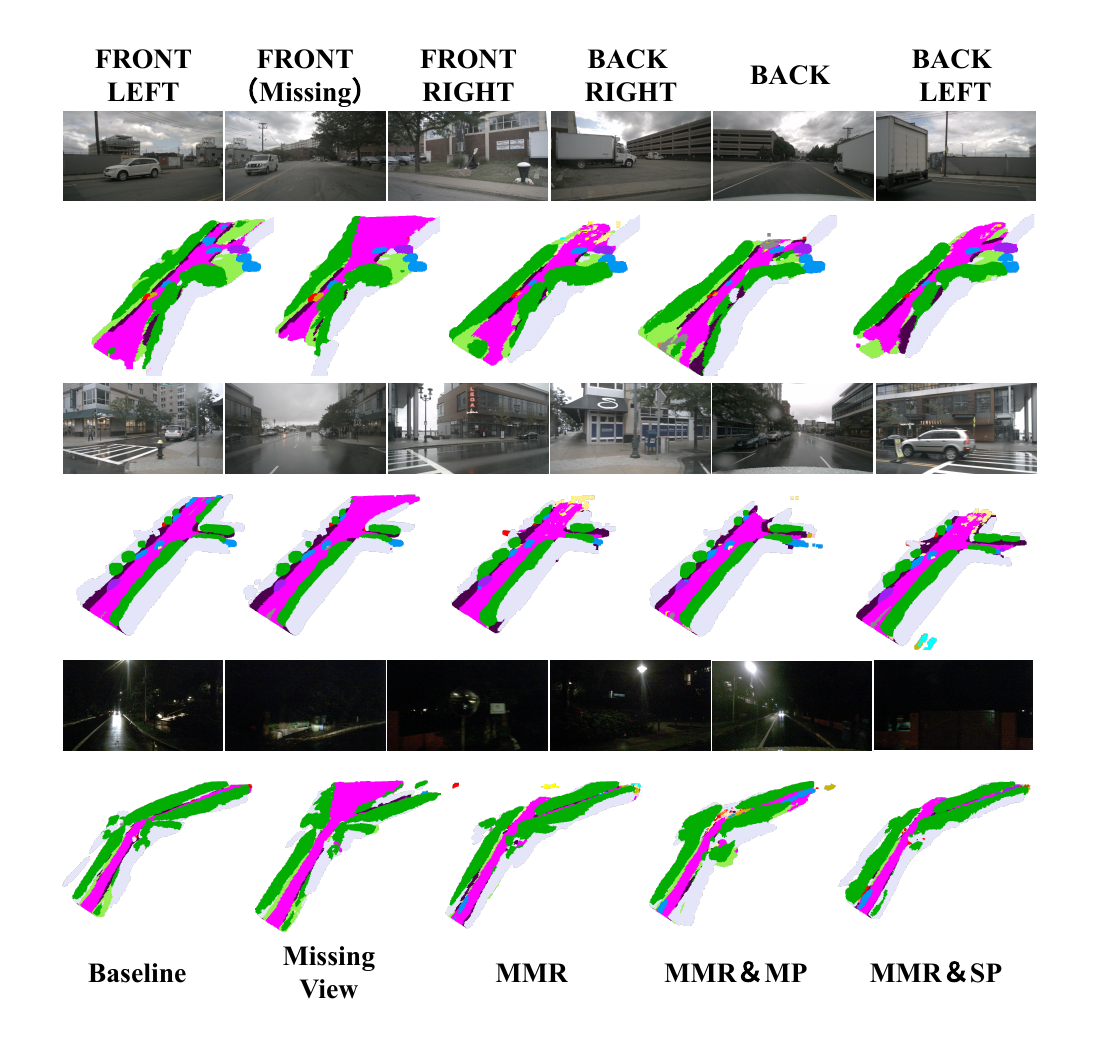}
    \vskip-1ex
    \caption{Visualizations on the nuScenes validation set~\cite{caesar2020nuscenes}. 
    Our method achieves promising results in the reconstruction of the missing $F$ view across various scenarios, even under weak lighting conditions. 
    }
    \label{fig:vis}
    \vskip-1ex
\end{figure}

\begin{figure}[!t]
    \centering
    \includegraphics[width=0.5\textwidth]{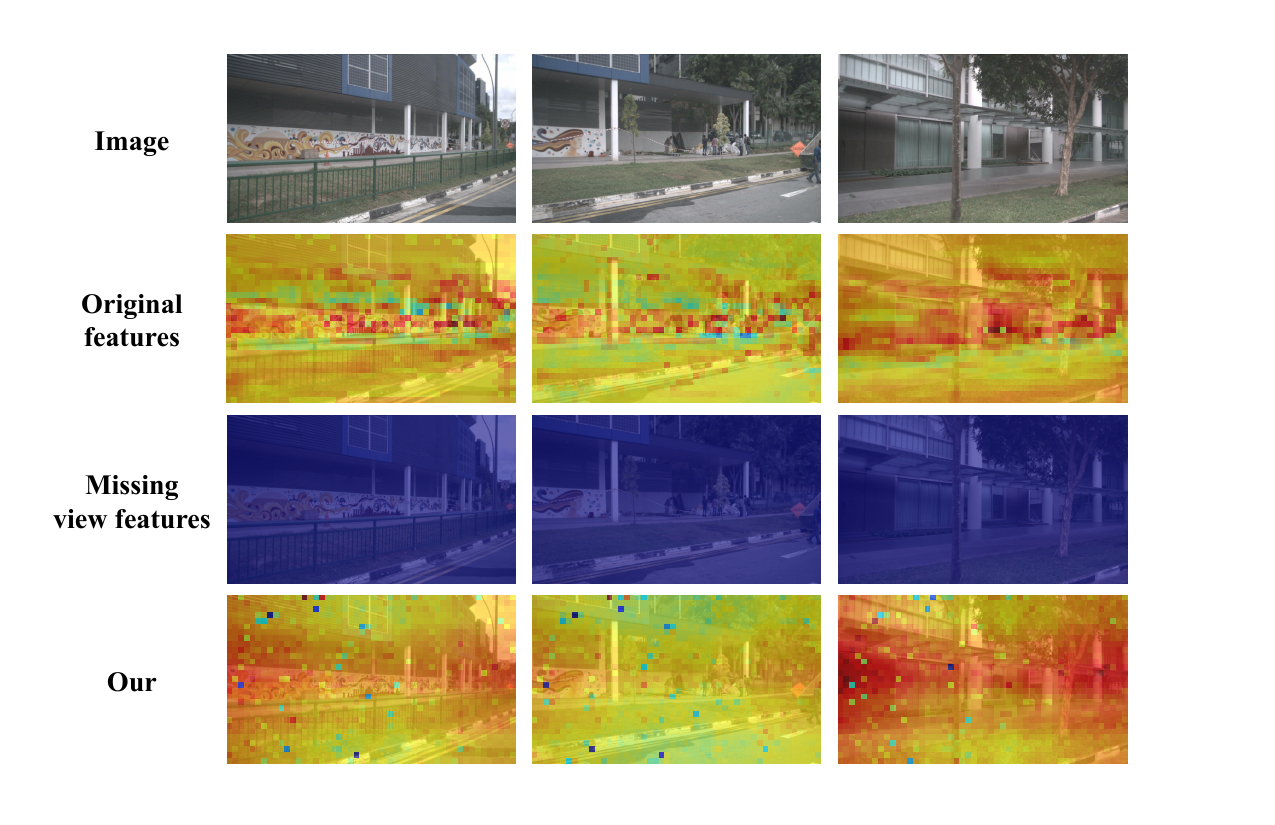}
    \vskip-1ex
    \caption{Visualization of feature maps. Our reconstructed features can largely replace the original features.
    }
    \label{fig:feat}
    \vskip-1ex
\end{figure}

In Figure~\ref{fig:vis}, we visualize the occupancy prediction results. In the ``missing FRONT view'' scenario, the baseline method fails to detect the road and vehicles in the front blind spot, producing fragmented geometry and missing semantics. In stark contrast, $M^2$-Occ successfully hallucinates the drivable surface and vehicle structures, closely matching the Ground Truth and demonstrating its ability to infer occluded regions from contextual cues. This qualitative result aligns with our quantitative improvements, where our method achieves substantial gains across most categories.

While our framework exhibits strong capability in reconstructing large-scale structures (\textit{e.g.}, roads, vehicles) under missing views, we acknowledge that fine-grained details of small objects (\textit{e.g.}, distant pedestrians) remain challenging to fully recover. Nevertheless, the overall perceptual quality is significantly enhanced: across diverse scenarios with missing front views, $M^2$-Occ consistently produces coherent scene layouts where the baseline collapses, confirming its effectiveness as a robust solution for real-world deployment where sensor failures are inevitable.

In Figure~\ref{fig:feat}, we present the original feature maps, the feature maps when a certain viewpoint is missing, and the reconstructed feature maps under several scenarios. 
Our reconstructed features can serve as a reasonable substitute for the original features, particularly for the detection of large objects, including drivable areas, various types of vehicles, and man-made structures. 
Meanwhile, the reconstruction of small objects and fine-grained textures remains challenging and requires further optimization in future work.

\subsection{Model Efficiency}

\begin{table}[!t]
    \centering
    \caption{Experimental results of computation overhead.}
    \label{tab:compute_overhead}
    \begin{tabular}{c|c|cc}
        \toprule
        \textbf{Method} & \textbf{Number of Missing Views} & \textbf{Latency(s)} & \textbf{Memory(G)} \\
        \midrule

        \ Baseline & - & 0.50 & 5.927 \\
        \midrule

        \multirow{5}{*}{Ours} & 1 & 0.77 & \multirow{5}{*}{6.077} \\
         & 2 & 0.91 & \\
         & 3 & 1.00 & \\
         & 4 & 1.11 & \\
         & 5 & 1.25 & \\
        \bottomrule
    \end{tabular}
    \vskip-1ex
\end{table}

We compared the inference time and memory usage of the baseline and our method, as shown in Table~\ref{tab:compute_overhead}. All experiments were conducted on an RTX A6000 GPU. Compared with the baseline, our method increases video memory consumption by only approximately $0.15$ GB (about $2.5\%$) while achieving significant performance gains.

The inference latency increases with the number of missing views, as the MMR module sequentially recovers features for each missing view using a transformer decoder. This trade-off introduces a computational overhead that is justified by the substantial improvement in perception reliability under sensor failure, making it suitable for safety-critical driving applications.

\section{Conclusion and Future Work} 
This work addresses semantic occupancy prediction under incomplete multi-camera observations, a practical yet underexplored challenge for autonomous driving. We propose $M^2$-Occ, which enhances robustness via feature-level reconstruction and class-level semantic regularization. On the SurroundOcc benchmark, $M^2$-Occ consistently restores geometric completeness under sensor failures. In safety-critical missing back-view scenarios, it recovers blind-spot structures, achieving $+4.36$ IoU and $+0.78$ mIoU gains over the baseline. As more cameras are removed, the robustness advantage grows, underscoring the value of spatial redundancy and semantic priors when visual evidence is sparse. Without adding sensors or modifying the backbone, $M^2$-Occ reduces reliance on fully functional surround-view systems, contributing to more reliable autonomous perception.

We acknowledge limitations in preserving fine-grained details for small objects under severe missing-view conditions. Future work will explore multi-resolution feature reconstruction, uncertainty-aware refinement, and temporal consistency to better recover small objects and mitigate the impact of instantaneous sensor failures.

{\small
\bibliographystyle{IEEEtran}
\bibliography{bib}
}

\end{document}